\documentclass{svproc}
\usepackage{url}
\usepackage{amssymb}
\usepackage{multirow}

\usepackage{xcolor}
\usepackage{xspace}

\usepackage{amsmath,amsfonts}
\usepackage{algorithmic}
\usepackage{algorithm}
\usepackage{caption}
\usepackage{subcaption}

\usepackage{graphicx}
\usepackage{cite}
\usepackage{tabularx} 
\usepackage[colorlinks=true,linkcolor=blue,citecolor=blue,urlcolor=blue]{hyperref}

\begin{document}
\mainmatter              % start of a contribution
\title{Performance Analysis of a Mass-Spring-Damper Deformable Linear Object Model in Robotic Simulation Frameworks}
\titlerunning{DLO simulation Comparison}  % abbreviated title (for running head)
\author{ Andrea Govoni\inst{1} \and   Nadia Zubair\inst{1} \and Simone Soprani\inst{1} 
\and Gianluca Palli\inst{1}}
\authorrunning{A. Govoni et al.} % abbreviated author list (for running head)
\tocauthor{Andrea Govoni, Nadia Zubair, Simone Soprani, Gianluca Palli}
\institute{$^1$Alma Mater Studiorum - Università di Bologna, Bologna 40126, IT,\\
\email{andrea.govoni11@unibo.it},
\email{nadia.zubair@studio.unibo.it},
\email{simone.soprani2@unibo.it},
\email{gianluca.palli@unibo.it}
\texttt{ }
}

\maketitle              % typeset the title of the contribution

\begin{abstract}

The modelling of Deformable Linear Objects (DLOs) such as cables, wires, and strings presents significant challenges due to their flexible and deformable nature. In robotics, accurately simulating the dynamic behavior of DLOs is essential for automating tasks like wire handling and assembly. The presented study is a preliminary analysis aimed at force data collection through domain randomization (DR) for training a robot in simulation, using a Mass-Spring-Damper (MSD) system as the reference model. The study aims to assess the impact of model parameter variations on DLO dynamics, using Isaac Sim and Gazebo to validate the applicability of DR technique in these scenarios.
Code available at https://github.com/GovoUnibo/cable$\_$model/settings

\end{abstract}
\section{Introduction}
\label{s:Introduction}
\begin{figure} [b]
\centering
    \begin{subfigure}[b]{0.5\textwidth}
        \centering    
        \includegraphics[width=0.8\linewidth]{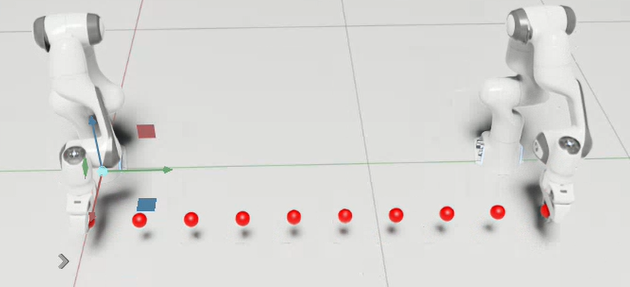}
        \caption{Isaac Sim Simulation Setup}
        \label{fig:IsaacSimSetup}
    \end{subfigure}
    \hfill 
    \begin{subfigure}[b]{0.45\textwidth}
        \centering   
        \includegraphics[width=0.78\linewidth]{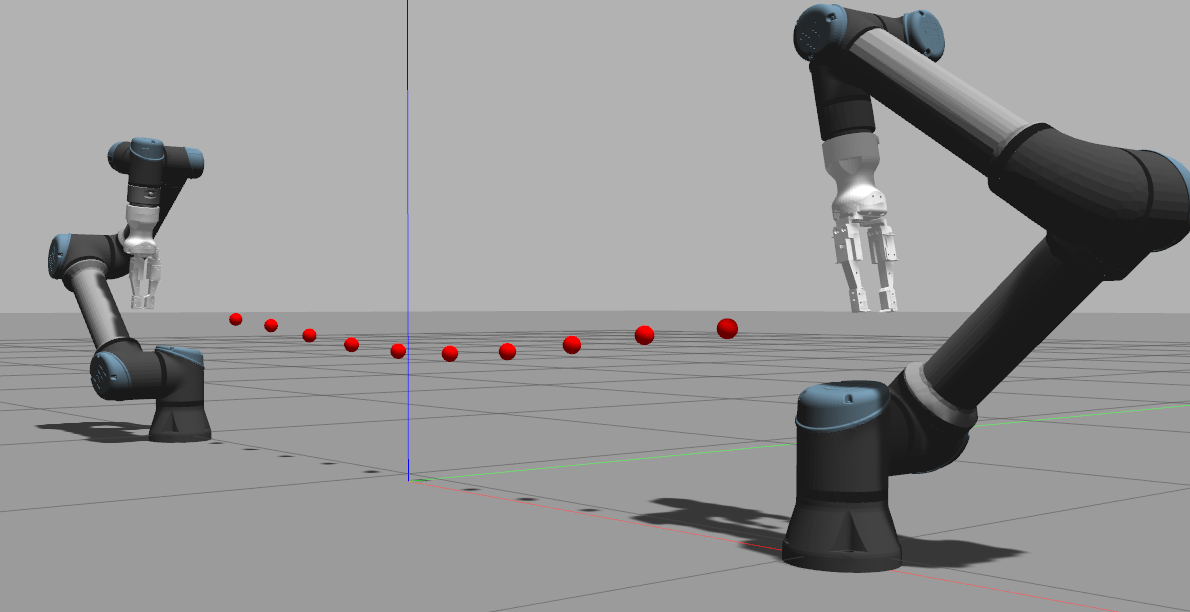}
        \caption{Gazebo Simulation Setup}
        \label{fig:GazeboSetup}
    \end{subfigure}
\caption{wire setups of Isaac Sim and Gazebo}
\label{fig:sim_setup_comparison}
\end{figure}
Nowadays there is an increasing demand for automation in industries that handle DLO, such as automotive wire harness installation \cite{wireharness}, cabling, and manipulation of flexible materials in manufacturing \cite{yang2022learning}. Manipulating DLOs presents a unique challenge due to their inherent unpredictability under varying forces and constraints, which complicates robotic control \cite{lv2022dynamic}. In these contexts, a thorough understanding of DLO behavior is essential, offering significant benefits, especially when simulation is used to optimize manipulation strategies.

Several different DLO model approaches are available, each with its own advantages and challenges \cite{DBLP:journals/corr/abs-2108-08935}, \cite{lv2020review}. For instance, the multibody approach \cite{rigid} and finite element methods \cite{KAUFMANN2009153} offer detailed simulations but come with high computational costs \cite{ac2024}. Other models, such as those based on dynamic splines \cite{splines}, \cite{article}, provide alternative methods but without a physical approach. While these models provide valuable insights into the behavior of DLOs, their effectiveness and applicability can vary depending on the intended use. 

In contemporary research, cable models are increasingly employed in simulations to rapidly generate datasets for training purposes. Our goal is to utilize the DLO models described in \cite{lv2017physically}, which discretize the DLO into point masses connected by dampers and springs. This MSD approach enables the collection of interaction force data alongside the cable’s position during manipulation with a dual robotic manipulator. These data serve as a foundation for refining control strategies, such as reinforcement learning (RL) \cite{SassanoRL}, to enhance bimanual manipulation tasks.

Achieving realistic DLO behavior through modeling is challenging. To address this, we plan to apply Domain Randomization (DR) in our simulations to improve the adaptability and performance of our control strategies in real-world scenarios \cite{DomainRandomization}. DR involves training the model with a wide range of variations in simulation parameters. However, before implementing this approach, it is crucial to examine how these variations impact the model's performance, particularly by identifying conditions that could lead to diminished effectiveness. Consequently, this work focuses on varying the parameters of the DLO model, with a specific emphasis on studying performance changes in robotic simulators.

The paper is organized as follows: in Sec. \ref{sec:eq_param} we describe the cable model with an emphasis on the parameters that influence its dynamics. Sec. \ref{sec:R&C} presents the results, iIllustrating how parameter variations impact the chosen Deformable Linear Object (DLO) model within the specific use case, presenting result validated using robotic simulators such as Isaac Sim and Gazebo.

\section{Simulation Setup Parametrs} \label{sec:eq_param}
\begin{figure}[!t] 
\centering
\includegraphics[width=0.5\columnwidth]{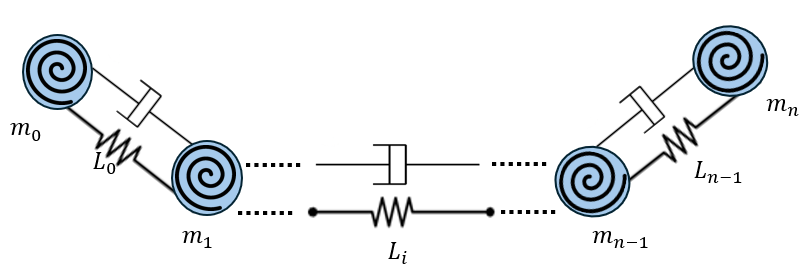}
\caption{Cable discretized in n masses ($m_i$) connected by n-1 Segments ($l_i$)}
\label{fig:cable_discretization}
\end{figure}

As shown in Fig. \ref{fig:sim_setup_comparison}, the DLO in simulation is represented as a discrete series of masses, each connected to adjacent masses through virtual springs (see Fig. \ref{fig:cable_discretization}). The behavior of mass $i$ is described by Newton's second law in Eq. \ref{eq:motion_eq}, which accounts for the elastic forces acting on it.

\begin{equation} \label{eq:motion_eq}
    m_i \frac{\partial^2 x_i}{\partial t^2} + k_d \frac{\partial x_i}{\partial t} = -\frac{\partial E_i}{\partial x_i} + F^e_i
\end{equation}
\(m_i\) represents the mass of the point, \(x_i\) is its position, \(k_d\) is the damping coefficient, \(E_i\) represents the total energy stored in the springs, and \(F^e_i\) are external forces such as gravity or interaction forces from the environment.

In this framework, we will focus on \textit{linear springs} and \textit{bending springs} indicated respectively as $l_i$ link in Fig. \ref{fig:cable_discretization} and spirals in Fig.\ref{fig:cable_discretization}. \textit{Twisting springs} are excluded from the simulation, since they are unnecessary for our DLO manipulation purposes. Further details can be found in \cite{lv2017physically}.

\subsection{Linear and Bending Springs in the Simulation} \label{formulas}

The energy stored in a linear spring ($E_i^s$) between two points is given by:

\begin{equation} \label{eq:linear_forces}
    \begin{aligned}
        E_i^s &= \frac{1}{2} k_s (l_i - l_{0i})^2 \\ 
        F_i^s &= -k_s (l_i - l_{0i}) \mathbf{u}
    \end{aligned}
\end{equation}

where, \( l_i \) is the current length between the mass points, \( l_0 \) is the initial, resting length of the segment, and \( \mathbf{u} \) is a unit vector in the direction of the spring’s force. The stiffness \( k_s \) is calculated as:

\begin{equation}\label{eq:linear_spring}
   k_s = \frac{E  A}{l_{0i}}
\end{equation} 

where, \(E\) is Young's modulus and \(A\) is the cross-sectional area of the wire. The force \( F_s \) from the spring is then used to update the positions of the connected mass points at each time step.

For bending, the energy associated with bending is given by:

\begin{equation}\label{eq:bending_forces}
    \begin{aligned}
        E_i^b &= \frac{1}{2} k_b \beta_i^2 \\ 
        F_i^b &= - \left( \frac{\partial E^b_{i-1}}{\partial x_{i}} + \frac{\partial E^b_i}{\partial x_i} + \frac{\partial E^b_{i+1}}{\partial x_{i}} \right)
    \end{aligned}
\end{equation}

where \( \beta_i \) is the angle between segments, calculated as:

\[
\beta_i = \arctan\left( \frac{ |\mathbf{l}_{i+1} \times \mathbf{l}_i|}{ \mathbf{l}_{i+1} \cdot \mathbf{l}_i } \right)
\]

Here, \( \mathbf{l}_i \) is the vector representing the current segment between two mass points, and \( \mathbf{l}_{i-1} \) and \( \mathbf{l}_{i+1} \) represent adjacent segments. The bending stiffness \( k_b \) is computed as:

\begin{equation}\label{eq:bending_spring}
    k_b = \frac{E I}{l_{0i}}
\end{equation}

where, \( I \) is the moment of inertia of the wire cross-section. These formulas are adapted for real-time force calculations among masses to update each simulation step. The computed forces are input for the simulator, which then uses integration, to update the DLO model’s velocities and positions.

\subsection{Damping and Integrator Step Problem} \label{sec:damping}
Choosing an appropriate damping coefficient (\(k_d\)) is challenging, as it must balance stability and responsiveness; a low $k_d$ can result in excessive oscillations, while a high $k_d$ may cause the system to become overdamped or even unstable.\\ 
Although using variable time steps would improve stability and accuracy in complex numerical simulations, simulators like Gazebo and Isaac Sim often rely on fixed step sizes, making it necessary to choose corresponding $k_d$ values.
To ensure system stability with a fixed time step, we apply the following helping condition found in  \cite{StabilityMSD}:
\begin{equation}
    k_d \leq \frac{|v m \Delta t + F|}{|v|}
    \label{eq:damping_factor}
\end{equation}
where \(v\) is the velocity, \(\Delta t\) is the  fixed for time step for integration to update the simulation, and \(F\) is the internal forces. \\

\section{Experiments \& Conclusion} \label{sec:R&C}
\begin{table}[!b]
\centering
\caption{Parameters used in the experiments} % Add a caption to the table
\label{tab:param} % Place the label after the caption
\begin{tabular}{|c|c|c|c|c|c|}
\hline
No. & E (MPa) & $m_i$ discretization & \(\Delta t\) Sampling Period (s) & Stability & Sec To Instability \\ \hline
1 & 12.6 & i=10 & 0.000005 & Stable & $\inf$ \\ \hline
2 & 526.0 & i=10 & 0.000005 & Stable & $\inf$ \\ \hline
2 & 1002.0 & i=6 & 0.000005 & Stable & $\inf$ \\ \hline
3 & 1002.6 & i=10 & 0.000005 & Unstable & 0.4 \\ \hline
4 & 1002.6 & i=10 & 0.0000001 & Stable & $\inf$ \\ \hline
\end{tabular}
\end{table}

\begin{figure}[!t] 
\centering
\includegraphics[width=0.65\linewidth]{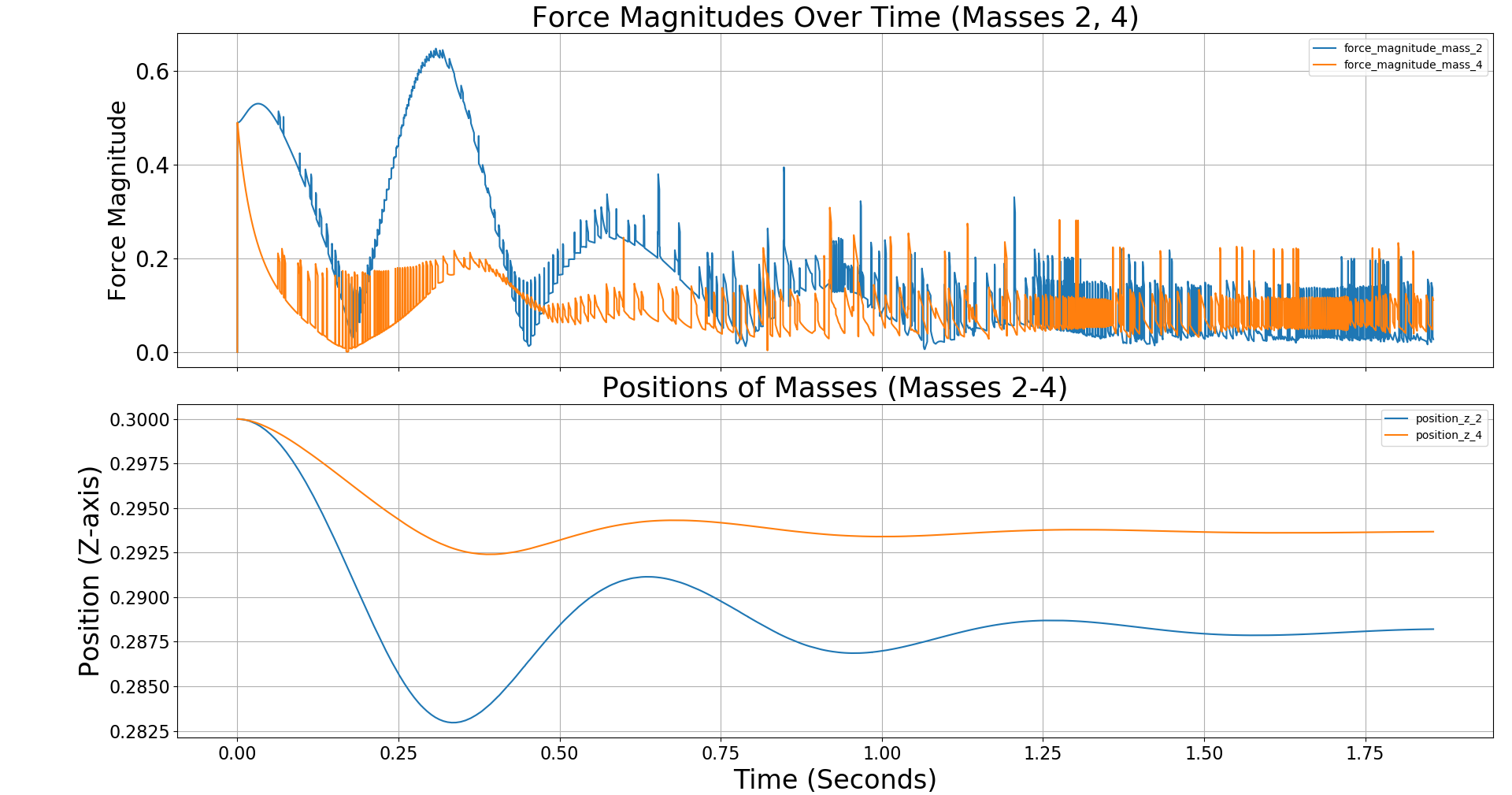}
\caption{Analysis of the motion along the z-axis of the masses 2,4, moving an extremity of 5 mm, using E=12.6MPa, mass discretization = 6 and, 0.000005 Sampling period}
\label{fig:cable_foces}
\end{figure}

\begin{figure}[!b] 
\centering
\includegraphics[width=0.7\linewidth]{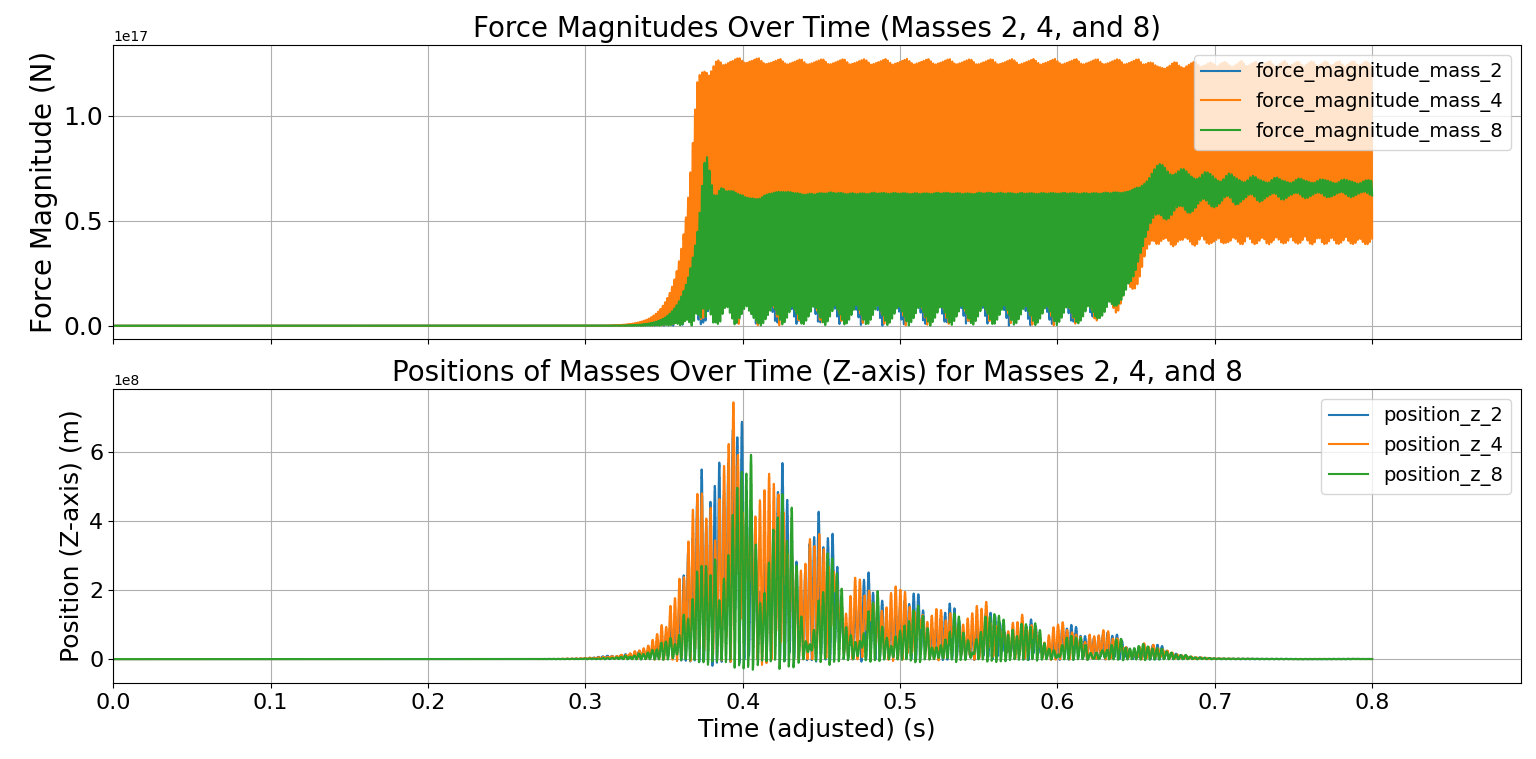}
\caption{Anlayzing masses 2,4,8, using E=1002.6MPa, mass discretization = 10 and, 0.00005 Sampling period lead to instability after 0.4 seconds}
\label{fig:unstable behavior}
\end{figure}

As previously mentioned in Sec. \ref{s:Introduction} accurately modeling a DLO using a fixed set of parameters is challenging due to the variability of real-world dynamics. To capture these complex behaviors, DR becomes essential for generating a dataset that enables algorithms to learn from a wide range of DLO scenarios in simulation. However, as described in Sec. \ref{sec:damping} it can be simple to see that varying parameters as mass or springs without careful adjustments of damping parameter ($k_d$) or sampling period ($\Delta t$) can lead to stability issues.

This section presents results from a specific use case, demonstrating that DR, requires a thoughtful approach, particularly for applications like this. We first implement the DLO model in both Isaac Sim and Gazebo simulators. Then, we start modifying parameters such as Young’s modulus and mass discretization, which affect the stiffness coefficients (Eq. \ref{eq:linear_spring} and Eq. \ref{eq:bending_spring}) to evaluate the model’s behavior. The case study implemented in the simulation is based on a DLO, whose initial conditions are described in Table 2 of \cite{lv2017physically}, where we explore the effects of varying Young’s modulus and mass discretization. To analyze the DLO’s stability under critical conditions, we created a controlled test environment. Initially, we fixed both ends of the cable parallel to the ground, allowing for an assessment of stability under static conditions. Following this, we introduced movement at one end of the cable to observe its dynamic response to displacement.

The results indicate similar DLO behavior in both simulators, achieved by adjusting the simulation step sizes to their respective nominal frequencies: 60 Hz for Isaac Sim and 1000 Hz for Gazebo. Although not presented here due to space constraints, the trends observed in both platforms reinforce the robustness of these findings. Additionally, due to space limitations, we focus on presenting only the graphs that highlight critical behaviors and stability challenges, omitting those that show stable configurations without issues.

As summarized in Table \ref{tab:param}, fixed step size lower values of Young's modulus produced as expected stable configurations, performing well under tested conditions. A critical aspect to highlight in Fig.\ref{fig:cable_foces} is, even when stability is achieved, randomly chosen parameters can induce oscillations in the forces experienced by the cable which during simulated manipulation will be transmitted to the simulated force/torque sensor. These fluctuations reduce the suitability of the data for force control learning, leading to low-quality datasets.

In conclusion we can claim that, while parameter randomization can improve learning capabilities of the robot, careful considerations are needed for DLO models. Randomizing parameters without caution may yield unreliable results: in the worst case seemingly stable simulation can still produce inaccurate behaviors (Fig. \ref{fig:cable_foces}). Future work should focus on developing techniques for effective RL + DR throught every physical parameter that the cable offers and validate the RL approach in real world scenario using \cite{caporali2023rt} as feedback control law. To mitigate instability issues or explore alternative simulators with more advanced numerical methods than those available in Gazebo or Isaac Sim for this approach.

% \bibliography{references}

\end{document}